\documentclass{article}

\usepackage[final]{neurips_wrl2023}

\usepackage[utf8]{inputenc} %
\usepackage[T1]{fontenc}    %
\usepackage{hyperref}       %
\usepackage{url}            %
\usepackage{booktabs}       %
\usepackage{amsfonts}       %
\usepackage{nicefrac}       %
\usepackage{microtype}      %
\usepackage{natbib}
\usepackage{graphicx}
\usepackage{float}
\usepackage{amsmath}
\usepackage{amssymb}
\usepackage{array}

\title{Language-Conditioned Semantic Search-Based Policy for Robotic Manipulation Tasks}

\author{
 Jannik Sheikh\\
  Bielefeld University\\
  Bielefeld, Germany \\
  \texttt{jsheikh@techfak.uni-bielefeld.de} \\
  \And
  Andrew Melnik \\
  Bielefeld University\\
  Bielefeld, Germany \\
\texttt{andrew.melnik.papers@gmail.com
} \\
  \AND
  Gora Chand Nandi \\
  Indian Institute of Information Technology\\
  Allahabad, India\\
  \texttt{gcnandi@iiita.ac.in}\\
  \And
  Robert Haschke \\
  Bielefeld University\\
  Bielefeld, Germany \\
  \texttt{rhaschke@techfak.uni-bielefeld.de} \\
}

\begin{document}

\maketitle

\begin{abstract}
Reinforcement learning and Imitation Learning approaches utilize policy learning strategies that are difficult to generalize well with just a few examples of a task. In this work, we propose a language-conditioned semantic search-based method to produce an \textit{online search-based policy} from the available demonstration dataset of state-action trajectories. Here we directly acquire actions from the most similar manipulation trajectories found in the dataset. Our approach surpasses the performance of the baselines on the CALVIN benchmark and exhibits strong zero-shot adaptation capabilities. This holds great potential for expanding the use of our \textit{online search-based policy} approach to tasks typically addressed by Imitation Learning or Reinforcement Learning-based policies. Project webpage: \href{https://j-sheikh.github.io/behavioral-search-policy}{https://j-sheikh.github.io/behavioral-search-policy}
\end{abstract}

\section{Introduction}

In recent years, the field of robotics has significantly evolved, with robots becoming more powerful, versatile, and interactive, due to progress in the field of natural language processing, computer vision \citep{rana2023contrastive}, reinforcement learning \citep{schilling2019approach,8675643,bach2020learn}, and imitation learning \citep{hussein2017imitation}.

\paragraph{Motivation} The ability of any agent to interact within an environment seamlessly depends largely on its capacity to collect, process, and understand data that are largely unstructured. This data forms the agent's perception and guides its decisions, actions, and reactions. 
Instead of the traditional approach of complex training of a policy to solve specific tasks, our work explores a framework for solving various robot manipulation tasks by using a semantic search-based approach to generate an \textit{online search-based policy}, inspired by the work of \cite{malato2023behavioral, beohar2022planning, beohar2022solving, rana2023contrastive}.

\begin{figure}
  \centering
  \includegraphics[width=0.8\textwidth]{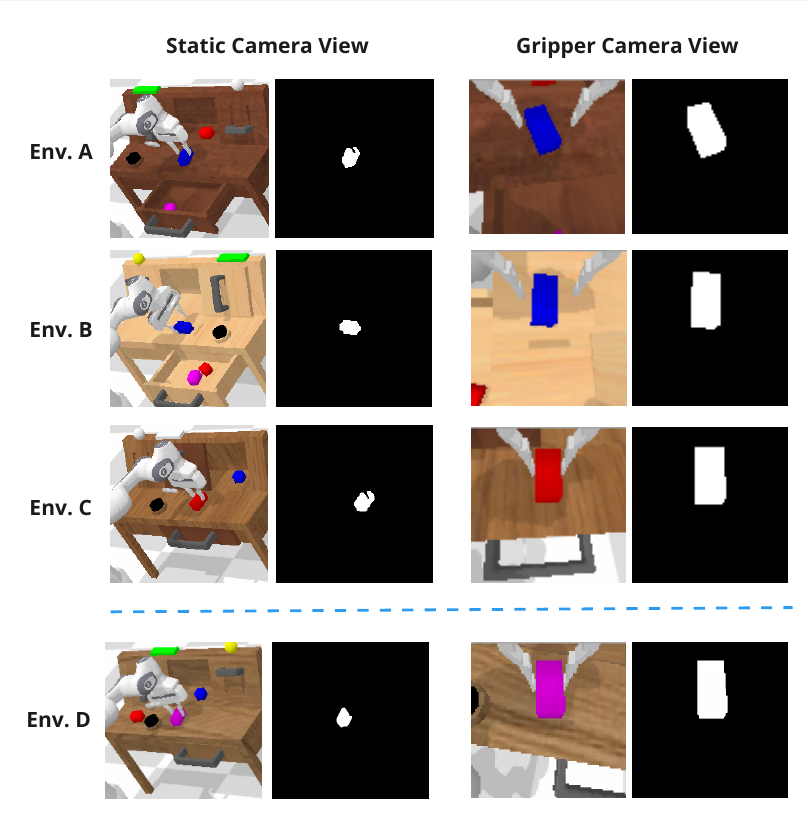}
  \caption{Overview of all four different environments in CALVIN. During inference, the \textit{Search-Based Policy} searches for the most similar state in environments A, B and C with respect to the current state from environment D.
  }
  \label{fig:calvinabcd}
\end{figure}

\section{Data}

The CALVIN benchmark \citep{mees2022calvin} contains four different tabletop environments (A, B, C, and D) as seen in Figure \ref{fig:calvinabcd}. Those environments always contain a desk with stationary and movable objects to interact with, whose initial positions vary over the environments. A drawer and sliding door can be opened and closed. A button toggles an LED light and a switch operates a light bulb. Further three different-sized, colored, and shaped blocks are somewhere located on the desk. A 7-DoF robot arm with a parallel gripper is used to interact with the environment.

The demonstration dataset of the benchmark was obtained from teleoperated "play" data, thus consisting of state $x_i$ and action $a_i$ pair trajectories $\tau$. Therefore $\tau$ contains the exact information of how the agent, controlled by a human, got from some initial state $x_0$ to a goal state $x_g$, for completing a task. This guarantees that the goal state is reachable from the initial state under the performed actions. This results in a dataset $D_{\text{play}} = \{\tau | \tau = \{(x_i, a_i)\}_{i=0}^{n} \text{ and } 0 \leq n \leq 64 \}$.

The authors labeled less than 1\% of the collected data, making it possible to identify trajectories that correspond to specific tasks. By annotating these trajectories, they made it possible to describe any of the trajectories, and thus the corresponding tasks, by natural language instructions.

\section{Method}

\begin{figure}
  \centering
  \includegraphics[width=1.0\textwidth]{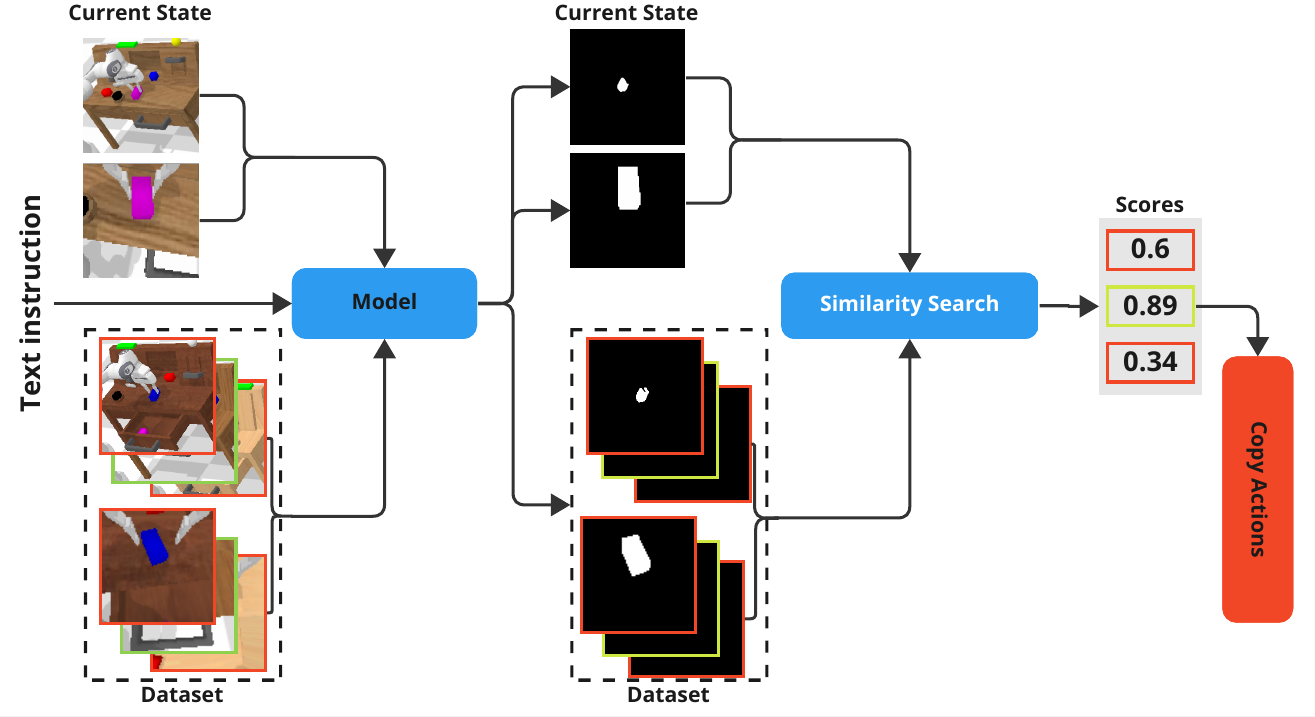}
  \caption{Overview of our framework. Given $x_t$, we obtain a binary mask of the object of interest in the static and gripper camera views and then compute $sim_{zs}$ to find the most similar state in dataset trajectories and start cloning the corresponding actions.}
  \label{fig:model}
\end{figure}

Instead of training a policy to solve tasks, we used search in the latent space of object shapes \citep{melnik2021critic, rothgaenger2023shape} to identify similar states in a demonstration dataset, and after finding similar representations for a given scene and text task, we clone the corresponding actions to solve the given tasks until the divergence threshold is exceeded between the current state and the selected trajectory.

\paragraph{Masking}

Since all objects over all the environments are of color, we first experimented with transforming the current state $x_t$ obtained by the static and the gripper camera views into latent spaces $z_{ts}$ and $z_{tg}$ by color-based and low-level feature-based segmentation methods. Concretely, we use the \textit{HSV (Hue, Saturation, and Value)} color space to detect and segment objects in our environments. We further enhanced our segmentation by adding positional filtering. Concretely, since there exist three different gray color handles we consider the position of the object in the image to help identify the correct one. This is easily applicable to the static camera view. The gripper camera added additional complexity because this camera is constantly in motion. Here we considered the surrounding pixels as well as the area and orientation of the object to make better and more accurate decisions. To be able to find the object of interest in our dataset, we apply a mapping to recognize the objects of interest in different environments based on the task description.

\paragraph{Search}
We obtain reference images $img_{ts}$ and $img_{tg}$ from the static and gripper camera view, capturing the current state $t$ in the environment. By passing $img_{ts}$ and $img_{tg}$ through our segmentation pipeline conditioned on text \( l \) we obtain two latent representations: \( z_{ts} \) and \( z_{tg} \). 

Each \(\tau \in D_{\text{play}|l}\) consists of a series of $x_{is}$ and $x_{ig}$, where $x_{is}$ refers to the RGB images of $x_i$ the static camera and $x_{ig}$ of the gripper camera. Both $x_{is}$ and $x_{ig}$ give a visual representation of the agent's progress toward the target object. Analogous to our approach described for the reference images, these sequences can be processed to obtain the corresponding latent representations $s_{is}$ and $s_{ig}$. 
By selecting images at every \( i^{\text{th}} \) step from \(\tau\), we further optimize for computational efficiency.
\paragraph{Similarity Measurement}
Given the reference latent representations \(z_{ts}\) and \(z_{tg}\) for state $t$, and the latent representations \(s_{is}\) and \(s_{ig}\)
from a trajectory \(\tau \in D_{\text{play}|l}\), we derive a weighted similarity coefficient as follows:

\begin{equation}
sim_{zs} = \alpha \cdot \text{score}(z_{tg}, s_{ig}) + (1 - \alpha) \cdot \text{score}(z_{ts}, s_{is})
\label{eq:sim_zs}
\end{equation}

Here, the \textit{score} is defined by the Dice coefficient:

\[ \frac{2 |A \cap B|}{|A| + |B|} = \frac{2 \text{TP}}{2 \text{TP} + \text{FP} + \text{FN}} \]

In addition, we scale the dice coefficient by a size coefficient. This coefficient serves to help identify latent representations containing objects of similar size, thus capturing the physical proximity of the robot arm to the objects. This primarily influences the relationship between $z_{tg}$ and $s_{ig}$, since $z_{ts}$ and $s_{is}$ are always captured from the same distance to the table. When objects in the binary masks $m_{tg}$ and $m_{ig}$ have nearly equivalent sizes, it indicates that the robot arm is at a similar distance from the objects in both scenarios. The size coefficient, \( size\_coef \), is then defined as the ratio between $z_{tg}$ and $s_{ig}$.

Finally, the weighted dice coefficient is calculated as:

\begin{equation}
    \text{weighted\_dice\_coef (\textit{score})} = \text{dice\_coef} \times \text{size\_coef}
    \label{eq:weighted_dice}
\end{equation}

\paragraph{Initiating the Search Process}

Given that the length of each trajectory is finite and contains at most 64 state-action pairs and over time the observed state will differ from the initial search and, consequently, from the trajectory we copy, we keep track of the standard deviation of $sim_{zs}$. 
After each execution of \(a_i \in \tau\), we collect the next state \(x_{i+1}\) from \(\tau\) and generate \(s_{i+1s}\) and \(s_{i+1g}\). Concurrently, we obtain the newly observable state in the environment \(x_{t}\), from which we derive \(z_{ts}\) and \(z_{tg}\). We then compute $sim_{zs}$ and store the results. If the change in the standard deviation over the last two steps exceeds a certain threshold \(T\) or if there are no actions left in the current pursued trajectory, we trigger a new search with the current state \(x_t\).

\paragraph{Switching Trajectories}
If the similarity score $sim_{zs}$ between \(z_{ts}\),\(z_{tg}\) and \(s_{is}\), \(s_{ig}\) is greater than the similarity value of the currently pursued trajectory, we switch to the \( i^{\text{th}} \) step of that new trajectory. Moreover, if there are no actions left in the current trajectory we pursue, we switch to \(\tau\) corresponding to the highest $sim_{zs}$ computed in the given step. 

\paragraph{Executing Actions}
Our action set has both absolute (\(a_{abs}\)) and relative (\(a_{rel}\)) actions. \(a_{abs}\) enable long-range movements that allow the agent to quickly reduce the distance to the target object. In contrast, \(a_{rel}\), allows finer movements that are important for local control and adjustment. 
Once \(z_{tg}\) contains non-zero values and our similarity score $sim_{zs}$ exceeds a certain threshold, the agent switches to \(a_{rel}\).

\section{Results}

CALVIN benchmark offers different evaluation environments.
The search-based policy uses the dataset that contains only three of the four CALVIN environments (A, B, and C), and is evaluated on the unseen environment D (see Figure \ref{fig:calvinabcd}).

\paragraph{Evaluation Settings}
We evaluated the agent's performance in two different ways (see Table \ref{tab:combined_results}):

\begin{enumerate}
    \item The agent performs each of the 34 tasks for 10 rollouts. Every evaluation starts by resetting the environment and agent to the initial state $x_0$ of an unseen demonstration.
    \item The agent is evaluated over 1000 individual task instructions. At the beginning of each task evaluation, the robot arm is placed in a neutral position and the environment is initialized. The agent's goal is to successfully perform the given task within a maximum of 360 steps. 
\end{enumerate}

There are two baseline models, MCIL \citep{lynch2021language} and HULC \citep{9849097}, although HULC is evaluated only in the second setting. 

\textbf{Hyperparamters}: For both evaluation settings we assign a value of 0.9 to $\alpha$ (Eq. \ref{eq:sim_zs}), emphasizing the latent representation of the gripper camera in our search. This choice is influenced by our intention to prioritize the gripper camera as soon as an object is detected in its view to allow precise maneuvers and finer adjustments. The static camera primarily helps in approaching the object and provides slight guidance in the further course.  In order to fasten the processing time of finding the most similar representation in the training data, we uniformly take only 8\% of the annotated data from each of the three environments A, B, and C, and search within this subset. Finally, we only consider every fourth image in a trajectory. We trigger a new search if the standard deviation of $sim_{zs}$ is greater than 0.03. If $z_g$ is empty, we set the threshold to 0.003 based on the Equation \ref{eq:sim_zs}.

\begin{table}[h]
    \centering
    \small
    \begin{tabular}{lcccl}
        \toprule
        \multicolumn{1}{c}{\textbf{Method}} & \textbf{Input} & \textbf{Success Rate First Setting} & \textbf{Success Rate Second Setting}\\
        \midrule
        Baseline & Static RGB \& Gripper RGB & 38\% & 30.4\% \\
        HULC & Static RGB \& Gripper RGB & & 41.8\%  \\
        \textbf{Ours} & Static RGB \& Gripper RGB &\textbf{61.4}\% & \textbf{57.2}\% \\
    \bottomrule
    \end{tabular}
    \vspace{0.15in}
    \caption{Combined results for the Zero-Shot Multi Environment in different evaluation settings.}
    \label{tab:combined_results}
\end{table}

\textbf{First evaluation setting)}:
As seen in Table \ref{tab:combined_results} our proposed method outperforms the current baseline by more than 23\% in the first setting. 

A detailed breakdown across all tasks can be found in Table \ref{tab:resultaltabletasks}. %

\begin{table}[htbp]
  \centering
  \begin{tabular}{|>{\raggedright\arraybackslash}p{3.5cm}|c|>{\raggedright\arraybackslash}p{3.5cm}|c|}
    \hline
    Task & Success Rate & Task & Success Rate \\
    \hline
    push pink block left & 100\%& rotate pink block left& 80\% \\
    push red block left & 100\%& rotate red block left& 90\% \\
    push blue block left & 70\%& rotate blue block left & 30\% \\
    push pink block right & 90\%  & rotate pink block right & 40\% \\
    push red block right& 20\% &  rotate red block right & 70\% \\
    push blue block right& 80\% & rotate blue block right & 40\% \\
    push into drawer& 0\% & unstack block& 70\% \\
    lift pink block drawer & 90\% & stack block& 0\% \\
    lift red block drawer& 70\% & turn on led & 90\% \\
    lift blue block drawer & 90\% & turn off led & 50\% \\
    lift pink block slider & 50\% & turn on lightbulb & 70\% \\
    lift red block slider& 20\% & turn off lightbulb & 80\% \\
    lift blue block slider & 10\% &  place in drawer & 100\% \\
    lift pink block table& 40\% & place in slider & 30\% \\
    lift red block table& 30\% & move slider right& 80\% \\
    lift blue block table & 50\% & move slider left & 70\% \\
    open drawer & 100\% & close drawer& 90\% \\
    \hline
  \end{tabular}
  \caption{Our results over all tasks in the first evaluation setting.}
  \label{tab:resultaltabletasks}
\end{table}

\textbf{Second evaluation setting}:
The results for the second evaluation setting are also shown in Table \ref{tab:combined_results}.

The agent completes 75\% of the tasks in 125 steps or fewer. It is important to emphasize that we set the maximum step size of our evaluation to 180. This decision arises from the observations of prior experiments, which indicate that the probability of failure is high if our strategy does not solve the task within this step range, and also to further reduce the computation time for each task.

\paragraph{Large-Scale Models}

Since we have natural language instructions that can serve as targets $x_g$, these instructions can be used to identify the relevant objects that we want to encapsulate in our latent space. With large language models, we can encode our task instructions and use the resulting latent representation to drive the segmentation process, which can also be replaced with an additional foundation model to identify the objects needed to solve the task.

\begin{figure}
  \centering
  \includegraphics[width=1.0\textwidth]{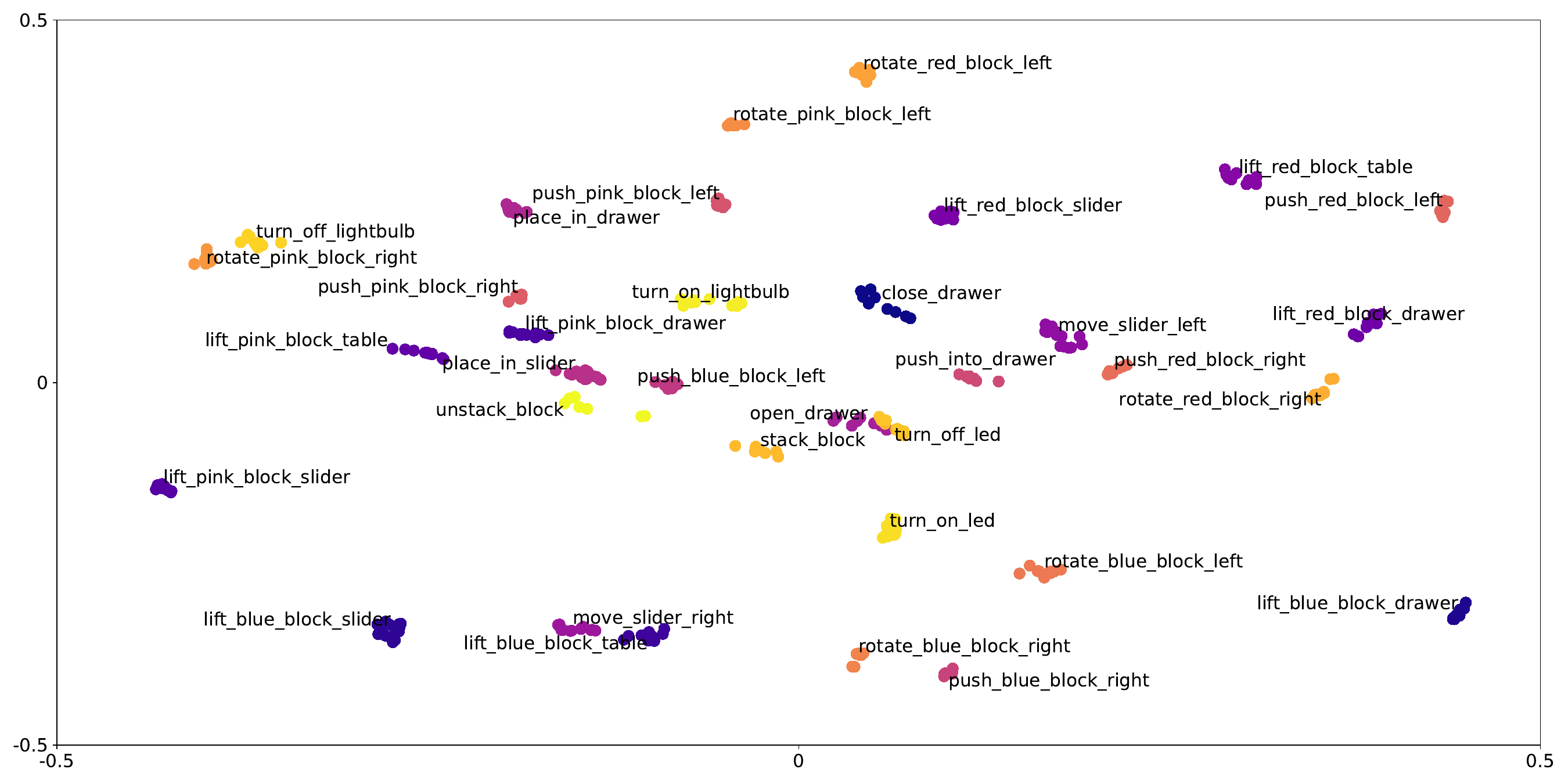}
      \caption{Visualization of the clustered natural language instructions with PCA in a 2D space. For clustering, we fit K-Means to the train embeddings of size \( \mathbb{R}^{768} \) generated by the fine-tuned GTE model \citep{li2023general} and set the number of clusters to $k$, where $k$ represents the total number of tasks, 34. Each data point represents a unique natural language instruction corresponding to a task, and the cluster labels denote the respective tasks. The plot shows the embeddings from our training data.}
    \label{fig:llmclustering}
\end{figure}

\begin{figure}
  \centering
  \includegraphics[width=1.0\textwidth]{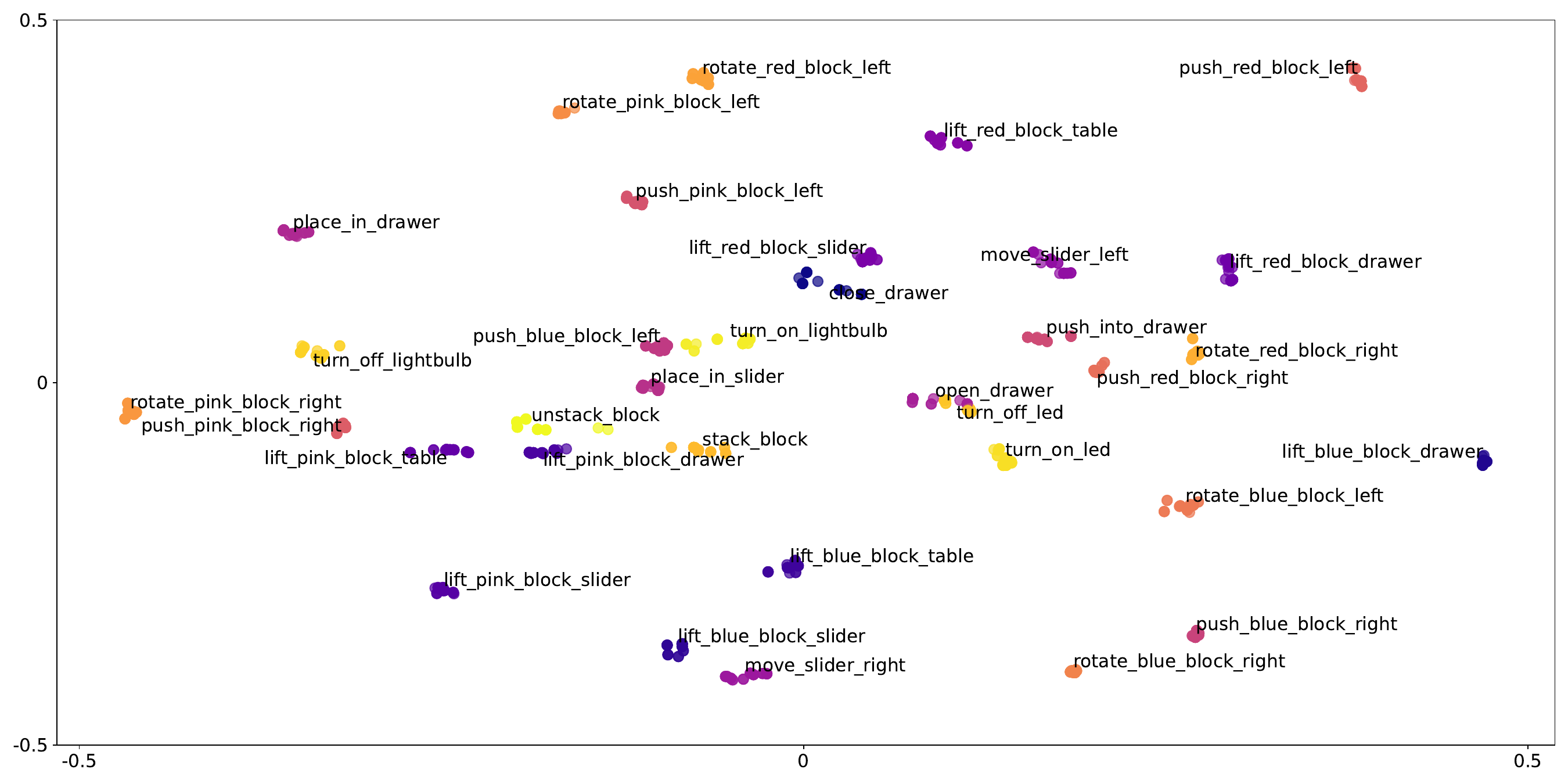}
      \caption{Visualization of clustered natural language test instructions with PCA in a 2D space. We generate the embeddings from the fine-tuned GTE model \citep{li2023general} and use the fitted K-Means algorithm to predict the clusters of the test embeddings of size \( \mathbb{R}^{768} \).}
    \label{fig:testclustering}
\end{figure}

We leverage the large pre-trained text embedding model \texttt{$GTE_{base}$} \citep{li2023general} which is based on the BERT framework \citep{devlin2019bert} and can be used for various downstream tasks. We further fine-tuned the model for three epochs using an A100-GPU using the task instructions of our training data. Afterward, we use the fine-tuned model to generate embeddings of size \( \mathbb{R}^{768} \) for our train and test instructions. Our results are visualized in Figure \ref{fig:llmclustering} for the train instructions and in Figure \ref{fig:testclustering} for the test instructions. In order to objectively measure the quality and separability of the clusters generated from our embeddings, we further report the Silhouette Score, Adjusted Rand Index (ARI), and Normalized Mutual Information (NMI) from scikit-learn \citep{scikit-learn}. The silhouette score ranges from -1 to 1, with a high value indicating that the embedding fits well with its own cluster and poorly with neighboring clusters. ARI measures the similarity between true labels and predicted labels and NMI measures the similarity between true labels and predicted labels. The clustering results of our embeddings are shown in Table \ref{tab:clustering_scores}. These scores indicate an almost perfect separation.

\begin{table}[h]
\centering
\begin{tabular}{|c|c|}
\hline
\textbf{Metric} & \textbf{Score} \\
\hline
Silhouette Score & 0.943 \\
\hline
Adjusted Rand Index & 1.0 \\
\hline
Normalized Mutual Information & 1.0 \\
\hline
\end{tabular}
\caption{Clustering Evaluation Metrics}
\label{tab:clustering_scores}
\end{table}

\section{Discussion}

Our work shows that using an \textit{online search-based policy} that exploits latent representations achieves notable success in solving a variety of robot manipulation tasks. By searching for similar latent representations in a demonstration dataset and mirroring the associated actions, our proposed method outperforms the current baseline models in both evaluation settings and generalizes for multi-environments. This highlights the compelling effectiveness of using a search-based policy within the latent space, a result consistent with the research of \cite{malato2022behavioral, malato2023behavioral} within the dynamic world of Minecraft.

The first evaluation setting provided insight into the overall effectiveness of our approach across all tasks and highlighted its consistent performance over numerous rollouts. The second evaluation setting demonstrates the robustness and efficiency of our method, as the agent navigated from a neutral position to the target object and then completed the task reasonably fast. This highlights the potential to address robot manipulation challenges without relying on complex reinforcement or imitation learning policies.

The performance differences between similar tasks (see Table \ref{tab:resultaltabletasks}) with differently sized blocks
indicate that the decision process for starting a new search and transitioning to an alternative trajectory requires more research to enable better precision and adaptive interactions. When interacting with the large (pink) block, the success rate across all corresponding tasks is almost 20\% higher than when interacting with the small (blue) block. Possible improvements could include the use of specific cost, exponential function, or other non-linear functions. Such approaches could offer advantages in effectively modeling the relationships between the latent representations of the static and gripper cameras and improve decision-making.

We further had to reduce the dataset due to the computing time when performing our search. The search process itself to find the most similar state in \(D_{\text{play}|l}\) is performed on the CPU and takes between 1.4 and 1.9 ms. Future research could investigate the effectiveness of initially generating a dataset of latent representations and then using clustering and indexing techniques to improve search speed. This approach provides flexibility to easily incorporate new latent representations and thus allows for new tasks to be performed.

Finally, foundation models seamlessly align with our suggested architecture (see Figure \ref{fig:model}). Models such as GLIP \citep{li2022grounded} and FastSAM \citep{zhao2023fast} show promising capabilities in generating latent representations of the object of interest conditioned by natural language. Further research is needed to evaluate such models in combination with our \textit{online search-based policy}.

\section{Conclusion}
In this work, we propose a method for solving various robot manipulation tasks using semantic search in the demonstration dataset and copying actions from the best-matching trajectory. We show that the proposed method generalizes for multi-environments.

\bibliography{bibliography}
\clearpage

\end{document}